\documentclass[sn-mathphys,Numbered]{sn-jnl}

\usepackage{graphicx}%
\usepackage{multirow}%
\usepackage{amsmath,amssymb,amsfonts}%
\usepackage{amsthm}%
\usepackage{mathrsfs}%
\usepackage[title]{appendix}%
\usepackage{xcolor}%
\usepackage{textcomp}%
\usepackage{manyfoot}%
\usepackage{booktabs}%
\usepackage{algorithm}%
\usepackage{algorithmicx}%
\usepackage{algpseudocode}%
\usepackage{listings}%
\usepackage{amsthm}
\usepackage{amsmath} 
\usepackage{amsfonts}
\usepackage{amssymb}
\usepackage{booktabs}

\theoremstyle{thmstyleone}%
\theoremstyle{thmstyletwo}%

\theoremstyle{thmstylethree}%

\usepackage{bbm} 
\usepackage[shortlabels]{enumitem}
\begin{document}

\title{On a joint simultaneous learning of relevant feature subsets and subspaces in regression-like problems}


\author{\fnm{Illia} \sur{Horenko}}\email{horenko@rptu.de}


\affil{\orgdiv{Chair for Mathematics of AI, Faculty of Mathematics}, \orgname{RPTU Kaiserslautern-Landau}, \orgaddress{\street{Gottlieb-Daimler-Str. 48}, \city{Kaiserslautern}, \postcode{67663}, \country{Germany}}\vspace{-0.5cm}}

\abstract{
We extend a recently introduced Entropy-Optimal Manifold Clustering (EOMC) to allow for a joint simultaneous identification of \emph{subsets and subspaces} of relevant features in nonstationary and nonlinear regression problems. It is shown that the proposed extension - that we coin as 
 Entropy-Optimal Manifold Regression (EOMR) - allows a  robust  learning with linearly-scaling iteration and memory complexities.  EOMR is compared  to the most complete set of state-of-the-art tools from the Artificial Intelligence (AI) and Machine Learning (ML) that is available to the author, on the very challenging problems from chaotic and fluid dynamics: (i) on predicting the Lorenz-96 systems dynamics in strongly- and very-strongly chaotic regimes (with forcing parameter being $F=8$ and $F=12$, respectively); and, (ii) on a data from the  Hasegawa-Wakatani model on the edge of the tokamak plasma. It is demonstrated that  the proposed benchmarks (i) and (ii), indeed, are the very challenging problems for the state of the art ML and AI tools - since  both the general-purpose gradient boosted random forests and deep neuronal networks, as well as  transformer-based AI tools like TabPFN v.03 (more spezialised for large-dimensional small data learning problems)  - result in orders of magnitude inferior root mean squared prediction errors,  and orders of magnitude larger model complexities, when compared to the EOMR. For a Hasegawa-Wakatani example, EOMR distills a very simple entropy-optimal and skilful description of the leading Essential Orthogonal Function (EOF) dynamics, given by linear, causal and weakly-stationary autoregressive process described by just 8 parameters.}
\maketitle

\section{Introduction}
\label{sec:introduction}
Data-driven approaches and tools from Artificial Intelligence (AI) and Machine Learning (ML) are gaining increasing popularity in modelling and prediction of chaotic systems \cite{ham2019deep,ramadevi2022chaotic,ghadami2022data,jiao2024ai,brenowitz2025practical,groom26a,groom26b}.  However, applications of common data-hungry ML and AI tools to such systems - especially in practical real-life situations - are characterized by a so-called \emph{small data learning challenge}, when the feature space dimension is relatively large, and the available statistics of systems observation is  relatively small, for example, due to the expense of the direct numerical simulation (e.g., in fluid mechanics), or due to the relative shortness of available observation history (e.g., in weather and climate research) \cite{Horenko_2020,espa_22,horenko_pnas_23,TabPFN,groom26a,groom26b}. Moreover, when applying very recent tools based on Large Language Models (LLMs), main bottleneck is represented by the unfavourable computational and memory scalings of transformer-based architectures - that scale quadratically both in the feature dimension and in the data statistics size, making training and applications of such tools very expensive for large realistic problems  \cite{TabPFN,bassetti25}.

A key to success in such situations is to identify or to learn the important feature dimensions that are most relevant for  the considered problem. Existent \emph{dimension reduction} and  \emph{feature extraction} methods can be subdivided into two groups: (i) into the methods identifying the relevant \emph{subsets} of original feature dimensions, and (ii) the methods that identify the relevant \emph{subspaces}: 
\begin{itemize}
\item {\bf Subset methods} aim at finding a relevant subset of original features, and include approaches, like the methods based on 1D statistical importances (e.g., based on filtering-out all of the feature dimensions below a certain  p-value treshold), the very popular Ridge- and Lasso-regression methods (using $l2$ and $l1$ parameter norms as regularizers \cite{tibshirani96}), or entropic learning methods using Shannon entropy as regularizers for probability distributions of model parameters    \cite{Horenko_2020,espa_22,horenko_pnas_23,bassetti25}. 
\item {\bf Subspace methods} aim at finding a linear or a nonlinear transformation of the original coordinate system in the feature space, for example, by means of an appropriate rotation and projection on a low-dimensional manifold, like in the Principal Component Analysis (PCA) \cite{Tenenbaum2000Global,Jolliffe2002PCA}, as well as in its numerous  linear and nonlinear extensions \cite{izenman1975reduced,horenko06,horenko08,vanderMaaten2008Visualizing,majda12,vanderMaaten2014Accelerating,McInnes2018,Healy2024UMAP,umap_interp,umap_kernel,horenko26}. 
\end{itemize}

To address numerical issues resulting from the polynomial scalability of  subspace methods - making them exceedingly expensive for large-dimensional feature spaces - another large group of methods aims at combining these two approaches into pipelines, for example, by first pre-selecting a potentially-relevant subset of original features (e.g., via the p-value tresholding), followed by an application of the subspace-based method, in the way how it is implemented in supervised PCA, sparse PCA, and in the related algorithms \cite{bair2006prediction,zou2006sparse,hastie2009elements,barshan2011supervised}.  

In the following, we will present an extension of the recently-introduced unsupervised and linearly-scalable Entropy-Optimal Manifold Clustering method (EOMC) for feature extraction beyond global linearity and stationarity assumptions - that belongs to a family of \emph{subspace methods}. We will demonstrate that this extension - that we will call Entropy-Optimal Manifold Regression (EOMR), preserves the central advantages of EOMC (like applicability to nonlinear and nonstationary problems, linear scalability of iteration and memory complexities, as well as the interpretable probabilistic formulation that allows finding entropy-optimal models). Moreover, we will show that beyond these advantages of EOMC, EOMR allows a \emph{supervised} identification of features relevant for linear and nonlinear regression and autoregression problems - by means of a \emph{simultaneous and joint learning} of relevant  feature subsets \emph{\bf and} subspaces, implemented as a numerical solution of the unified, probabilistic, and entropy-regularized optimization problem. After analyzing the numerical properties of the proposed solution, and discussing the way of hyperparameter selection and relations to existing approaches, we exemplify the application of EOMR to two challenging problems: to Lorenz-96 model in strongly- and very-strongly-chaotic regimes, as well as to the  Hasegawa-Wakatani model from Magnetohydrodynamics.     

\section{Methods}
\label{sec:methods}
\subsection{Mathematical formulation of the Entropy-Optimal Manifold Regression (EOMR)}
\label{sec:m_formulation}
Let $X\in\mathbb{R}^{D,T}$ be a given $(D\times T)$-dimensional real-valued data matrix, with every column  $X(:,t)$ representing a  $D$-dimensional vector of feature values for a data instance with an index $t$, where $t=1,\dots,T$ ($:$ denotes a column-extraction operation).   Let $Y\in\mathbb{R}^{1,T}$ be a given series of values, for which we want to learn a functional relation that for any given instance $t$ should map  a vector $X(:,t)$ to a scalar value $Y(t)$\footnote{Without a loose of generalaty, we consider a scalar valued regression problem, where dimensionality of the values $Y(t)$ is equal to 1. Please note that all of the following derivations can be straightforwardly extended to multiple dimensions of the target variable $Y$. The reason is that in the case of $m>1$ dimensions, the optimization formulation provided below will be equivalent to $m$ independent optimization problems that can be solved separately.}.   
 Let there be $K$ "local" mappings from $X(:,t)$ to  $Y(t)$, each of them is characterized by  the orthogonal $d$-dimensional linear manifold projector $\mathcal{T}_k\in\mathbb{R}^{D,d}$, $k=1,\dots,K$, by $K$ of $1$-by-$(d+1)$-dimensional vectors of regression coefficients $\tilde\theta_k=\left[\theta^0_k,  \theta_k\right]$, and by a joint diagonal matrix $W = \text{diag}(w_1, \dots, w_D)$ with $w_i \ge 0$ and $\sum_{i=1}^Dw_i=1$, with a diagonal containing the probabilities that a particular feature dimension $i$ belongs to a relevant subset of features:
 
 \begin{eqnarray}
\label{eq:mod}
Y(t)&=&\sum_{k=1}^K\gamma(k,t)\left(\theta^0_k +\theta_k\mathcal{T}_k^{\dagger}WX(:,t) \right)+\delta_t,
\end{eqnarray}
where $\delta_t$ is the independent identically distributed scalar-valued random process with expectation zero, and $\gamma(k,t)\ge 0$, $\sum_{k=1}^K\gamma(k,t)=1$ for all $t$ being the probabilities that $Y(t)$ is generated by the "local" model $k$. In another words, $Y(t)$ is predicted as the expectation over $K$ subset-reduced (multiplication with $W$) and subspace-reduced (multiplication with $\mathcal{T}_k$, where $\dagger$ is the conjugate-transpose operation) $(d+1)$-dimensional linear regressions ($d<<D$), where expectation is taken with respect to the (a priori unknown) and time-dependent model probability distributions $\gamma(:,t)$.
  
Assuming in addition that the rows of feature matrix X are not perfectly correlated with each other, from the Gauss-Markov theorem \cite{gaussmarkov49}  it follows that the best unbiased estimator of model (\ref{eq:mod}) is provided by the minimum of the following least-squares functional:
 \begin{eqnarray}
\label{eq:LS}
\mathcal{L}_{ls}=\sum_{t=1}^T\left(Y(t)-\sum_{k=1}^K\gamma(k,t)\left(\theta^0_k +\theta_k\mathcal{T}_k^{\dagger}WX(:,t)\right) \right)^2.
\end{eqnarray}
Then, using strict convexity of the squared Euclidean norm and deploying the Jensen-inequality, we get the upper bound of (\ref{eq:LS}):
 \begin{eqnarray}
\label{eq:LS}
\mathcal{L}_{ls}\le\mathcal{L}_{J}=\sum_{t,k=1}^{T,K}\gamma(k,t)\left(Y(t)-\theta^0_k -\theta_k\mathcal{T}_k^{\dagger}WX(:,t)\right)^2.
\end{eqnarray}
 Please note that $\mathcal{L}_{ls}\equiv\mathcal{L}_{J}$ when $\gamma$ contains only zeros and ones. 
 
 Moreover, let $\mu_k\in\mathbb{R}^{D,1}$ for all $k$ from $1$ to $K$ be the centroid vectors in Euclidean space. Then, following the strategy recently proposed for Entropy-Optimal Manifold Clustering (EOMC) \cite{horenko26}, we add to $\mathcal{L}_{J}$ (multiplied with a non-negative hyperparameter $\epsilon_R$) a metrization term $\gamma(k,t) (X(:,t) - \mu_k)^\dagger W(X(:,t) - \mu_k)$\footnote{This metrization term is two-fold important: (i) as in EOMC, it removes the non-zero kernel from the manifold projection, and (ii) it will become indispensable when making predictions for the test data where $Y(t)$ is unknown a priori, since it becomes essential in determining the $\gamma(:,t)$ -and, hence, which regression should attain which weight in the model (\ref{eq:mod}).} , as well as the two regularization terms $\epsilon_\gamma\phi_1(\gamma(:,t))$ and $\epsilon_W\phi_2(W)$ with $\epsilon_\gamma,\epsilon_W\ge 0$. This results in the Enropy-Optimal Manifold Regression (EOMR) learning  of the least-biased version for model (\ref{eq:mod}), given the fixed data $X$ and $Y$, and, for the fixed selected values of hyperparameters $d,K,\epsilon_R,\epsilon_\gamma,\epsilon_W$ - formulated and implemented as the numerical solution for the following constrained optimization problem: 
\begin{eqnarray}
\label{eq:eomc}
 &&\left\{W,\gamma^*,\mu^*_1,\mathcal{T}^*_1,\tilde\theta^*_1,\dots,\mu^*_K,\mathcal{T}^*_K,\tilde\theta^*_K\right\}=\arg\min\mathcal{L}^{\textrm{EOMR}},\\
\label{eq:eomc2}
\mathcal{L}^{\textrm{EOMR}}&=&\frac{1}{T}\sum_{k,t=1}^{K,T}\gamma(k,t)\left[\|X(:,t) - \mu_k\|_W^2 +\epsilon_R\left(Y(t)-\theta^0_k -\theta_k\mathcal{T}_k^{\dagger}WX(:,t)\right)^2 +\epsilon_\gamma\phi_1(\gamma(:,t))\right]+\epsilon_W\phi_2(W),\nonumber\\
\label{eq:eomc3}
\textrm{s.t.}&&\mathcal{T}_k^\dagger\mathcal{T}_k=I_d, \\
\label{eq:eomc4}
&&\gamma(k,t)\geq0,\quad\textrm{and }\sum_{k=1}^K\gamma(k,t)=1,\quad \forall t,k,\\
\label{eq:eomc5}
&&W = \text{diag}(w_1, \dots, w_D),\quad w_i\geq0,\quad\textrm{and }\sum_{i=1}^Dw_i=1,\quad \forall i.
\end{eqnarray}
In the same way as it was done in EOMC \cite{horenko26} and in the other entropic AI methods, to achieve the entropy-optimal (i.e., least-biased) learning of probability distributions $\gamma$, a very good choice for $\phi_1(\gamma(:,t))$ would be  $\phi_1(\gamma(:,t))\equiv\log_K\left(\gamma(:,t)\right)$, which, after a multiplication with $\gamma(:,t)$, will result in a term that maximizes Shannon entropy - and, for all other variables being fixed, $\mathcal{L}^{\textrm{EOMR}}$ will have a unique and analytically-computable minimizer for $\gamma$,  given by a softmax function \cite{espa_22,horenko26}.

A good choice for  $\phi_2(\gamma(:,t))$ would be  to use the Fast Entropy Approximation function $\phi_2(W)\equiv\sum_{i=1}^DW_{ii}\left(\frac{0.6648}{W_{ii}+0.2086}-0.5754W_{ii}\right)+ 0.0206$. This choice has two main reasons: (i) $W$ appears inside of the squared Euclidean norm, resulting in the regularized Quadratic Programming (QP) problem - that, in contrast to the regularized linear problem for $\gamma$ - does not have an analytic solution for $W$ (at least, no analytic solution to the author's knowledge); and (ii)  FEA was recently shown to provide a very close, numerically-cheap, smoothly-differentiable and property-preserving approximation of the Shannon entropy, allowing for orders of magnitude faster and more sparse feature extraction in regression problems, when compared to common sparsification tools like Lasso-regression \cite{horenko26b}
\subsection{Numerical solution of  EOMR problem}
As in the case of EOMC and other entropic AI algorithms, mathematical structure of the optimization problem (\ref{eq:eomc}-\ref{eq:eomc5}) allows deploying subspace-iteration solution, i.e., after selecting hyperparameter values $d,K,\epsilon_R,\epsilon_\gamma,\epsilon_W$ and initial values of EOMR optimization variables $W,\gamma^*,\mu^*_1,\mathcal{T}^*_1,\tilde\theta^*_1,\dots,\mu^*_K,\mathcal{T}^*_K,\tilde\theta^*_K$, one can iteratively go through all of these EOMR variables, and solve the problem  (\ref{eq:eomc}-\ref{eq:eomc5}) for one of these variables at a time, while keeping all other variables frozen. It appears, that for the three of these five variable types (for $\gamma^*,\mu^*_1,\tilde\theta^*_1,\dots,\mu^*_K,\tilde\theta^*_K$) there exist cheap analytical solutions of the problem, that can be computed with the linear memory and time complexity scalings, whereas for the two remaining variable types (for $W,\mathcal{T}^*_1,\dots,\mathcal{T}^*_K$) we propose two linearly scalable numerical algorithms.   

\paragraph{Step 1: Minimization w.r.t. the manifold projectors $\mathcal{T}_k$ }
Freezing all of the  EOMR variables except of a $\mathcal{T}_k$  for a fixed $k=1,\dots,K$, we isolate the terms in $\mathcal{L}^{\textrm{EOMR}}$ that depend explicitly on $\mathcal{T}_k$. This yields a subproblem:
\begin{equation}
\min_{\mathcal{T}_k \in \text{St}(d,D)} f(\mathcal{T}_k) = \epsilon_R \sum_{t=1}^T \gamma(k,t) \left((Y(t) - \theta^0_k) - \theta_k \mathcal{T}_k^\dagger W X(:,t)\right)^2,
\end{equation}
where $\text{St}(d,D)$ is the so-called Stiefel-manifold \cite{edelman1998geometry,absil2009optimization}, defined by the quadratic equality constraint (\ref{eq:eomc3}).   
Let $z(t) = Y(t) - \theta^0_k$. Expanding the quadratic term gives:
\begin{equation}
f(\mathcal{T}_k) = \epsilon_R \sum_{t=1}^T \gamma(k,t) z(t)^2 - 2\epsilon_R \sum_{t=1}^T \gamma(k,t) z(t) \theta_k \mathcal{T}_k^\dagger W X(:,t) + \epsilon_R \sum_{t=1}^T \gamma(k,t) \left(\theta_k \mathcal{T}_k^\dagger W X(:,t)\right)^2.
\end{equation}
Using the cyclic property of the matrix trace ($\operatorname{tr}$), the linear interaction term can be rewritten as:
\begin{equation}
\sum_{t=1}^T \gamma(k,t)  z(t) \theta_k \mathcal{T}_k^\dagger W X(:,t) = \sum_{t=1}^T \operatorname{tr}\left( \gamma(k,t)  z(t) \mathcal{T}_k^\dagger W X(:,t) \theta_k \right) = \operatorname{tr}\left( \mathcal{T}_k^\dagger \left[ \sum_{t=1}^T \gamma(k,t)  z(t) W X(:,t) \theta_k \right] \right).
\end{equation}
We define the Euclidean gradient of the objective function with respect to $\mathcal{T}_k$ by differentiating $f(\mathcal{T}_k)$ directly:
\begin{equation}
\nabla_{\mathcal{T}_k} f = -2\epsilon_R \sum_{t=1}^T \gamma(k,t) \left(Y(t) - \theta^0_k - \theta_k \mathcal{T}_k^\dagger W X(:,t)\right) W X(:,t) \theta_k.
\end{equation}
To minimize $f(\mathcal{T}_k)$ while adhering to the Stiefel manifold, we project the Euclidean gradient $\nabla_{\mathcal{T}_k} f$ onto the tangent space $\mathbf{T}_{\mathcal{T}_k} \text{St}(d,D)$. The orthogonal projection operator yields the Riemannian gradient $\operatorname{grad} f(\mathcal{T}_k)$~\cite{edelman1998geometry}:
\begin{equation}
\operatorname{grad} f(\mathcal{T}_k) = \nabla_{\mathcal{T}_k} f - \mathcal{T}_k \left( \frac{\mathcal{T}_k^\dagger (\nabla_{\mathcal{T}_k} f) + (\nabla_{\mathcal{T}_k} f)^\dagger \mathcal{T}_k}{2} \right).
\end{equation}
A descent step is taken along the negative Riemannian gradient direction $\mathcal{V} = -\operatorname{grad} f(\mathcal{T}_k)$. To project this tangent vector back onto the manifold surface, we can deploy a QR-decomposition-based retraction mapping $\mathcal{R}_{\mathcal{T}_k}(\alpha \mathcal{V})$, where $\alpha > 0$ is a step size determined dynamically by an Armijo backtracking line search to ensure sufficient decrease~\cite{absil2009optimization}:
\begin{equation}
\mathcal{T}_{k,\text{cand}}(\alpha) = \mathcal{T}_k + \alpha \mathcal{V}, \quad \text{qrf}(\mathcal{T}_{k,\text{cand}}(\alpha)) = \mathcal{Q}\mathcal{R} \implies \mathcal{T}_{k,\text{new}} = \mathcal{Q} \cdot \text{diag}(\text{sign}(\text{diag}(\mathcal{R}))).
\end{equation}
\paragraph{Step 2: Minimization w.r.t. the centroid vectors $\mu_k$ }
Isolating the variance-like regularization block that contains $\mu_k$, we get:
\begin{equation}
\min_{\mu_k} \sum_{t=1}^T \gamma(k,t) ( X(:,t) - \mu_k)^ \dagger W ( X(:,t) - \mu_k).
\end{equation}
Taking the vector derivative with respect to $\mu_k$ and setting it to zero we get:
\begin{equation}
\nabla_{\mu_k} \mathcal{L}^{\textrm{EOMR}} = -2 \sum_{t=1}^T \gamma(k,t) W ( X(:,t) - \mu_k) = -2 W \left( \sum_{t=1}^T \gamma(k,t)  X(:,t) - \left(\sum_{t=1}^T \gamma(k,t)\right) \mu \right) = 0.
\end{equation}

Since $W = \text{diag}(W_{11}, \dots, W_{D,D})$, this decouples into $D$ independent scalar equations:
\begin{equation}
W_{ii} \left( \sum_{t=1}^T \gamma(k,t) X_{i,t} - \left(\sum_{t=1}^T \gamma(k,t)\right) \mu_i \right) = 0, \quad \forall i = 1, \dots, D.
\end{equation}

For any feature dimension where $W_{ii} > 0$, dividing by $W_{ii}$ yields the exact sample-weighted mean. For uninformative dimensions, where $W_{ii} = 0$, the coordinate has no effect on the functional value. Thus, the global minimizer simplifies to the stable sample-weighted mean vector across all dimensions:\begin{equation}
\mu_k = \frac{\sum_{t=1}^T \gamma(k,t)  X(:,t)}{\sum_{t=1}^T \gamma(k,t)}.
\end{equation}
\paragraph{Step 3: Minimization w.r.t. the reduced regression coefficients vectors $\tilde\theta_k=\left[\theta^0_k,  \theta_k\right]$ }
When all other variables except of $\tilde\theta_k$ are fixed, EOMR problem reduces to a linear least-squares problem. Let us define the augmented subspace feature vector $\tilde{x}_t \in \mathbb{R}^{(1+d) \times 1}$ as:
\begin{equation}
\tilde{x}_t = \begin{bmatrix} 1 \\ \mathcal{T}_k^ \dagger W  X(:,t) \end{bmatrix}
\end{equation}
The regression subproblem becomes:
\begin{equation}
\min_{\tilde{\theta}_k} \sum_{t=1}^T \gamma(k,t) \left( Y(t) - \tilde{\theta}_k \tilde{x}_t\right)^2.
\end{equation}
Taking the partial derivative with respect to $\tilde{\theta}_k$ and setting it to zero yields the standard regularized normal equations:
\begin{equation}
\label{eq:normal_eqs}
\left( \sum_{t=1}^T \gamma(k,t) \tilde{x}_t \tilde{x}_t^ \dagger\right) \tilde{\theta}_k^ \dagger = \sum_{t=1}^T \gamma(k,t)  Y(t) \tilde{x}_t
\end{equation}
Transposing this linear system gives the explicit global coordinator update for the row vector:
\begin{equation}
\tilde{\theta}_k = \left( \sum_{t=1}^T \gamma(k,t)  Y(t) \tilde{x}_t^ \dagger \right) \left( \sum_{t=1}^T \gamma(k,t) \tilde{x}_t \tilde{x}_t^ \dagger \right)^{-1}
\end{equation}
\paragraph{Step 4: Minimization w.r.t. the probability distributions $\gamma$ }
Applying the Euler-Lagrange principle with respect to $\gamma$ in (\ref{eq:eomc}-\ref{eq:eomc5}), and following  the same way as in the EOMC and other entropic AI methods, one obtains a following unique analytical solution of  (\ref{eq:eomc}-\ref{eq:eomc5}) :
\begin{eqnarray}
\label{eq:gamma1}
\gamma^*(k,t) &=&\frac{\exp\left(-\epsilon_\gamma^{-1}\left(g(k,t)-g(j^*(t),t)\right)\right)}{\sum_{k=1}^K\exp\left(-\epsilon_\gamma^{-1}\left(g(k,t)-g(j^*(t),t)\right)\right)},
\end{eqnarray}
 where $g(k,t)=\|X(:,t) - \mu_k\|_W^2 +\epsilon_R\left(Y(t)-\theta^0_k -\theta_k\mathcal{T}_k^{\dagger}WX(:,t)\right)^2$ and $j^*(t)=\text{argmin}_{k}g(k,t)$.
\paragraph{Step 5: Minimization w.r.t. the subset feature probability matrix $W$ }
Since $\phi_2(W)$ in (\ref{eq:eomc}-\ref{eq:eomc5})  was selected as a Fast Entropy Approximation (FEA), we can directly use the efficient and linearly-scalable Sequential Projected Gradient algorithm proposed in \cite{horenko26b} to find the optimizer with respect to $W$. 
\paragraph{Total iteration complexity and memory scaling of the subspace algorithm}
Summarizing the leading order scalings for the five steps described above, we obtain $\mathcal{O}(KDT+KDdT + Kd^2T + Kd^3)$ for iteration run time complexity, and $\mathcal{O}(KDT + KDd + KdT + Kd^2)$ as the memory complexity. Hence, if $d<<D$, both memory and iteration costs scale in the same order as the scaling of computationally very cheap clustering methods like K-means.  Steps 1 to 5 are repeated iteratively, and it is straightforward to validate that this procedure would result in the monotonic decrease of the functional value $\mathcal{L}^{\textrm{EOMR}}$,  until, at some iteration $(I)$, decrease of $\mathcal{L}^{\textrm{EOMR}}$ is less then some predefined tolerance threshold {\bf tol}. Total run time complexity in the leading order is then  $\mathcal{O}(KDTI + KDdI + KdTI + Kd^2I)$

\subsection{Selection of EOMR hyper-parameters $d,K,\epsilon_R,\epsilon_\gamma,\epsilon_W$}
\label{sec:hyper}
Since  (\ref{eq:eomc}-\ref{eq:eomc5}) is a supervised learning problem, we can directly apply most of the standard hyper-parameter selection routines from ML and AI, like cross-validation and Bayesian hyper-parameter tuning \cite{wu19}. Hereby, one first defines some reasonable ranges for the hyperparameters, and then subdivides the data $X$ and $Y$ into training, validation and testing subsets. For each of the particular values of the hyperpaparameters - chosen from the predefined ranges - one iteratively minimizes (\ref{eq:eomc}-\ref{eq:eomc5}) as described in the previous chapter, until the predefined tolerance threshold {\bf tol} is reached. Then, one evaluates the performance of the identified model on the validation data that was not used in training - and selects the hyperparameter combination with a best validation performance. Finally, the performance of the model with the best validation performance is further evaluated on the test data, that was not used neither in training nor in validation. In the following examples we will use this test performance to compare EOMR with other ML and AI tools, using the same cross-validation train/validate/test data splits.
   
\subsection{Relation to Existing Algorithms}
EOMR derived here is closely linked to several dimensionality reduction and regression methods in the machine learning literature - that can be considered as special asymptotic cases of EOMR:
\begin{enumerate}
\item \textbf{Reduced-Rank Regression (RRR) and Supervised PCA (SPCA):} When $K=1$, $\epsilon_R \to \infty$, $\epsilon_W \to \infty$,  and $\epsilon_\gamma = 0$, the problem approaches classical Reduced-Rank Regression (which fits a multivariate linear regression model under rank constraints~\cite{izenman1975reduced}), and the standard SPCA, that selects features and projects data based on maximizing a dependency metric (such as Hilbert-Schmidt Independence Criterion) between the data $X$ and the targets $Y$ ~\cite{bair2006prediction, barshan2011supervised}. 
\item \textbf{entropy-optimal Sparse Probabilistic Approximation (eSPA+):} When $\epsilon_R =0$,  and $\epsilon_\gamma = 0$, problem (\ref{eq:eomc}-\ref{eq:eomc5})  becomes equivalent to eSPA+  \cite{Horenko_2020,espa_22}. 
\end{enumerate}

\section{Application examples}
\label{sec:results}
\begin{figure}[h!]
 \centering
        \includegraphics[clip,  width=1.1\textwidth]{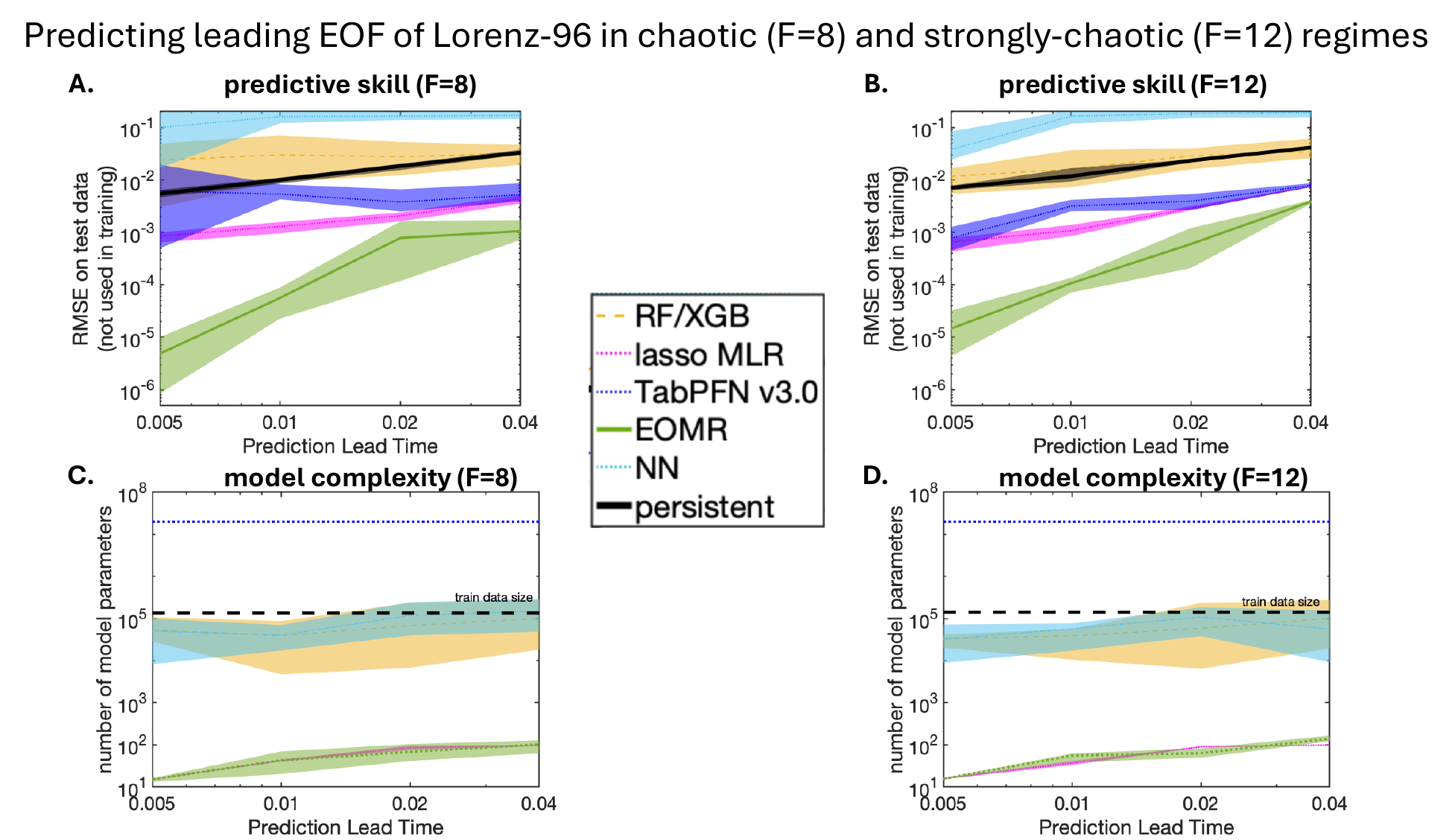}
        \includegraphics[clip,  width=1.1\textwidth]{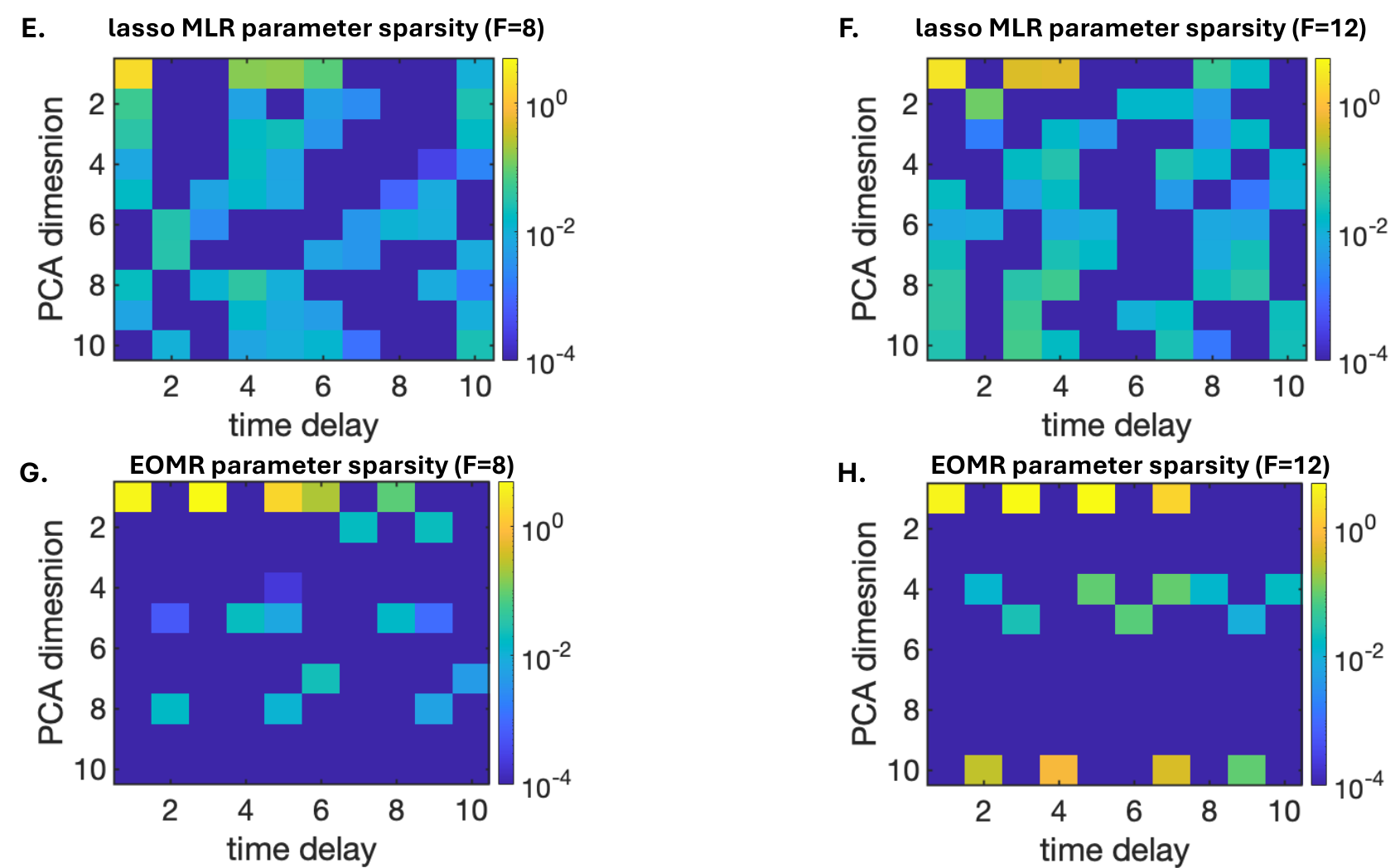}
 \caption{Analysis results for the L96 in the two chaotic regimes ($F=8$, left column, and $F=12$, right column). Explanations can be found in the Sec.~\ref{sec:lorenz}.}
\end{figure}
Below, we present application examples, and compare EOMR results to such AI tools as: (i) to the gradient-boosted random forests and XGBoost (abbreviated as RF/XGB, provided within a functionality of the Matlab function \emph{fitrensemble()}, with the automated adaptive hyperparameter tuning); (ii) to the Lasso-regularized sparse multilinear regression (abbreviated as lasso MLR, provided within a functionality of the Matlab function \emph{lasso()}, with the automated hyperparameter tuning in a range of $\left[10^{-10},10^{2}\right]$ for the regularization parameter);  to the Deep Neuronal Networks (abbreviated as NN, provided within  the Matlab Deep Learning Toolbox, with various network architectures, and with the numbers of hidden neurons ranging from 1 to 200, and the number of hidden layers ranging from 1 to 3);  and transformer-based TabPFN v3.0 model - a hyperparameter free LLM, specially designed for small data learning problems \cite{TabPFN}. For each of these models, we followed the same procedure as for EOMR (described in the Sec.~\ref{sec:hyper}),  and used the same cross-validation data splits to extract the best performing representatives from each model class, to be further compared to each other and to EOMR on the test data - that was held-out of training and validation steps. We also added comparisons to the so-called "persistent" prediction, when the next value of $Y(t)$ is taken to be the same as its previous value.
 \subsection{Analysis of data from Lorenz-96 in strongly-chaotic and very strongly-chaotic regimes}
 \label{sec:lorenz}
\paragraph{Model description} The Lorenz-96 (L96) model was introduced by Edward Lorenz in a 1996 paper (published later in 2005) as a  "toy model" of the Earth's atmosphere for studying the fundamental issues of predictability in chaotic dynamics in spatially extended systems \cite{Lorenz1996,Lorenz2005}. It mimics aspects of the mid-latitude atmosphere's non-linear dynamics, such as advection, dissipation, and external forcing, within a computationally cheap, periodic one-dimensional domain (a latitude circle). The L96 model is widely used today as a benchmark problem for data assimilation techniques, ensemble forecasting methods, and studies on the general nature of chaos \cite{Lorenz1998,anderson2001,houtekamer2005,bocquet2020}.

The L96 Type 1 represents a finite difference approximation of a partial differential equation describing a simplified 1D turbulence. Model consists of a system of $N$ coupled ordinary differential equations (ODEs), describing the time evolution of a single scalar atmospheric quantity $x_j$ at $N$ equally spaced grid points around a latitude circle:
\begin{equation}
\frac{dx_j}{dt} = (x_{j+1} - x_{j-2})x_{j-1} - x_j + F \quad \text{for } j = 1, \dots, N
\label{eq:L96}
\end{equation}
Periodic boundary conditions are assumed, such that indices are taken modulo $N$ (i.e., $x_{j+N} = x_j$ and $x_{j-N} = x_j$).
Variables and terms in (\ref{eq:L96}) have the following meaning:
\begin{itemize}
    \item $x_j$: The value of the atmospheric quantity (e.g., temperature, vorticity) at the $j$-th grid point.
    \item $N$: The total number of grid points in the system (system size). Common values in literature are $N=40$.
    \item $t$: Time.
    \item $F$: A positive, constant external forcing parameter that drives the system.
    \item $(x_{j+1} - x_{j-2})x_{j-1}$: The non-linear advection term, which conserves energy in the absence of forcing and damping.
    \item $-x_j$: A linear damping (dissipation) term.
\end{itemize}


Behaviour of the L96 model changes significantly with the forcing parameter $F$. For small values of $F$ (e.g., $F < 1$), the system exhibits periodic or steady-state dynamics. As $F$ increases, the system undergoes bifurcations and transitions into chaotic regimes. A commonly studied value is $F=8$, which produces robust chaotic behaviour used frequently as a standard benchmark in predictability studies. For regimes where $F \geq 7$ (which includes $F=8$ and $F=12$ investigated below), the system is considered to be in a strong or fully turbulent chaotic state \cite{Lorenz2005}. According to Andrew J. Majda (Courant Institute, deceased in 2021), development of scalable methods capable of robust predictions of L96 in these chaotic regimes represents a "800-pound gorilla" in the area of chaotic systems. 

\paragraph{Application of EOMR to L96 output data} 
In the following, we will use the common literature setting for $N=40$, and generate long time series $x\in\mathcal{R}^{40\times2000}$ of L96 for the two forcing regimes $F=8,12$,  with $T=10$ (corresponding to 50 Earth days after rescaling of L96-units) and time step $\tau=0.005$ (corresponding to 36 Earth minutes after rescaling of L96-untis) . Then, we perform the PCA transformation of $x$\footnote{In fluid mechanics and geosciences, PCA modes are called Essential Orthogonal Functions (EOFs), that is why we use this abbreviation in the Figures.},
 and choose as target $Y(t)$ the values of the dominant PCA mode (i.e., a mode with the largest variance), and choose as $X(t)$ the lag delayed embedding of the ten dominant  PCA modes of $x$ at times $\left(t-\text{lag}-n\tau\right)$, where variable $\text{lag}$ is varying between 0.005 and 0.04, and time delay index $n$ ranges between 0 and 9. This results in a feature space dimension $D=100$ (see Fig.~1).      


 In each of the forcing regimes we deploy the cross-validation procedure described above, and train EOMR models with hyperparameters sampled randomly from broad ranges of values: $d$ from $\left\{1,2,\dots,7\right\}$,  $K$ from $\in\left\{1,2,3\right\}$,  $\epsilon_R$ from the interval $\left[1,100\right]$,  $\epsilon_\gamma$ from the interval $\left[10^{-8},10^{-2}\right]$,  $\epsilon_W$ from the interval $\left[10^{-8},10^{2}\right]$. 
 
 As can be seen from the Figs.~1A-1B, for all of the considered prediction lag times, both Deep Learning NN and RF/XGB perform worse then the trivial "persistent" predictor, when the next value of $Y(t)$ is taken to be the same as its previous value (solid black lines in Fig.~1). And, this is despite of the quite extensive hyperparameter tuning for NN and RF/XGB.  Both TabPFN and lasso MLR perform close to each other, and around 7 times better than the "persistent" predictor. EOMR is the winner in prediction skill, achieving root mean squared error (RMSE) of around $7\times10^{-6}$ for a lag time $0.005$ in L96 time units (36 Earth minutes after rescaling),  with around 300 times smaller RMSE when compared to the next competitor lasso MLR, and approximately 2'100 smaller then the error of the trivial "persistent" predictor. Surprisingly, both RMSE and complexity (Fig.~1C and 1D) of the EOMR models do not change noticeably, when increasing F from 8 to 12 and going from strongly-chaotic to very-strongly chaotic L96 regime.  Complexities - i.e., the total numbers of tuneable model parameters that have to learned from the data - is close to the size of the training data - indicating that both of these models really struggle to learn, and just memorize the training data instead.   In contrast,  EOMR learns very sparse models (Figs.~1G-1H)  - even more sparse than the lasso MLR (Figs.~1E-1F, one of the most widely-used sparsification tools) - with just around a hundred of tuneable parameters that should be learned from data. 
 
 When making predictions for the new data points $X(:,t)$ with the piecewise-linear regression model (\ref{eq:mod}), it is sufficient to keep $K$ of $D$-dimensional vectors $\beta_k=\theta_k\mathcal{T}_k^{\dagger}W$. Then, making a prediction would mean computing $K$ scalar products $\beta_kX(:,t)$, and, since it requires only very cheap elementary operations like multiplication and addition - and none of the much more expensive operations like divisions - this can be done extremely efficiently, even on the common hardware architectures. For example, evaluating one single EOMR prediction in this L96 example required only around $10^{-8}$ seconds on the commodity Mac LapTop. In contrast, TabPFN v3.0 required around $10^{-2}$ seconds for the same data instances - due to the quadratic (both in $D$ and in $T$) scaling of transformer-based LLMs \cite{TabPFN}, and their huge size.      
 
\subsection{Analysis of data from modified Hasegawa-Wakatani (mHW) model of drift-wave turbulence in the edge of a tokamak plasma.}
 \label{sec:mHW}
 One-dimensional L96-model is considered by many to provide only a very limited, and simplified description of turbulent atmospheric dynamics - and not a model of a "real" turbulence behaviour. To check if the findings from previous Section are induced by this oversimplification - or if they are also reflecting intrinsic properties of "real" turbulent systems, next we will consider an application of EOMC to the output of a more realistic model from Magnetohydrodynamics (MHD).  
We will take the Hasegawa-Wakatani model - a seminal two-field fluid description model of drift-wave turbulence in the edge of a tokamak plasma \cite{hasegawa83,wakatani84,gottwald04}. It reduces the complex MHD equations to the evolution of density fluctuations (\(n\)) and electrostatic potential (\(\phi \)) fields. The modified Hasegawa-Wakatani (mHW) model describes the evolution of electrostatic potential $\phi$ and density fluctuations $n$ in a two-dimensional slab geometry. In the following application example, we will deploy the mHW model version introduced by Numata et al. \cite{numata2007}, involving the resistive coupling term acting only on non-zonal fluctuations. 
\begin{figure}[h!]
 \centering
        \includegraphics[clip,  width=1.1\textwidth]{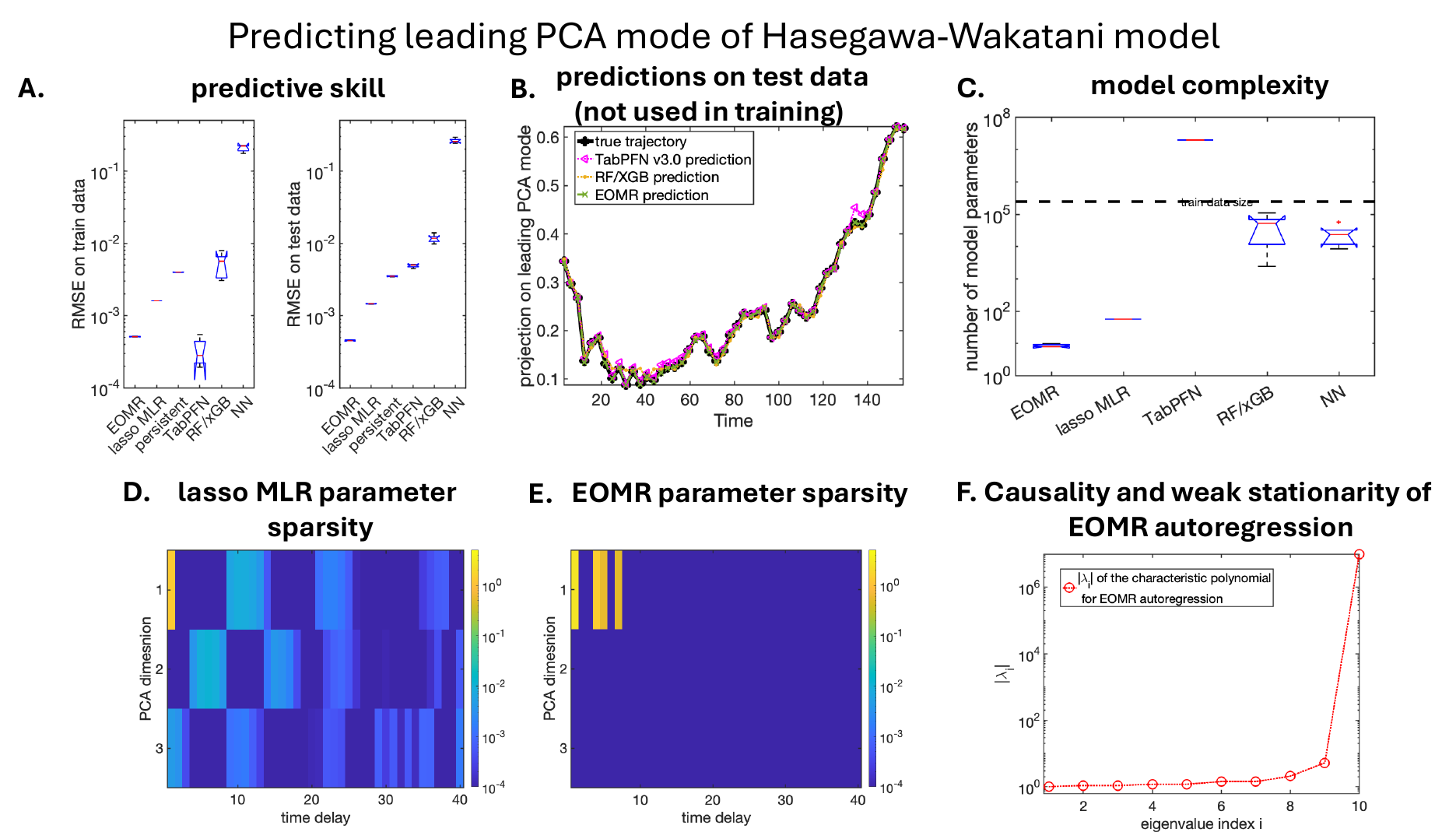}
 \caption{Analysis results for the modified Hasegawa-Wakatani model. Explanations can be found in the Sec.~\ref{sec:mHW}.}
\end{figure}

\paragraph{Governing Equations} 
Let the zonal average of a field $f$ be defined as $\langle f \rangle = \frac{1}{L_y} \int f \, dy$, and the non-zonal fluctuation as $\tilde{f} = f - \langle f \rangle$. The mHW equations are:
\begin{align}
    \frac{\partial \zeta}{\partial t} + \{\phi, \zeta\} &= \alpha (\tilde{\phi} - \tilde{n}) - \mu \nabla^4 \zeta, \label{eqn:vorticity} \\
    \frac{\partial n}{\partial t} + \{\phi, n\} + \kappa \frac{\partial \phi}{\partial y} &= \alpha (\tilde{\phi} - \tilde{n}) - \mu \nabla^4 n, \label{eqn:density}
\end{align}
where $\zeta = \nabla^2 \phi$ is the ion vorticity and $\{a, b\} = \partial_x a \partial_y b - \partial_y a \partial_x b$ is the Poisson bracket representing $\mathbf{E} \times \mathbf{B}$ advection.
Variables and parameters of mHW model equations \cite{numata2007} have the following meaning:
\begin{itemize}
\item    $\phi$: electrostatic potential;
\item     $n$: electron density fluctuations;
\item     $\zeta$: ion vorticity ($\nabla^2 \phi$);
\item     $\alpha$: adiabaticity parameter (resistive coupling);
\item     $\kappa$: background density gradient scale length ($-\partial_x \ln n_0$);
\item     $\mu$: dissipation/viscosity coefficient.
\end{itemize}
\paragraph{Data generation}
To generate the time series for  analysis and comparison, we use the MATLAB code by Jean-Christophe Nave and Denis St-Onge, available at \url{https://github.com/DenSto/HWE_solver}.  We use the same settings for all of the mWH model parameters as in this code, with the only change being a slightly reduced grid size ($64\times 64$ grid points). We generate a time series on the interval $\left[600,3000\right]$ (skipping the outputs between $t=0$ and $t=600$, when the model  "burns-in" and the transient states die-out), with constant time intervals $\delta t=0.3125$ (which, after rescaling from internal model units to seconds results in a time interval of $3.16\cdot 10^{-7}$ seconds, or $0.316$ microseconds)\footnote{Light in vacuum would travel 94.73 meters during this time interval.}.  

We follow the same protocol as in the previous example, and perform the PCA/EOF transformation of  the obtained simulation time series of ion vorticities and electron density fluctuations.
 We choose as target $Y(t)$ the values of the dominant PCA mode (i.e., a mode with the largest variance), and choose as $X(t)$ the lag delayed embedding of the three dominant  PCA modes  at times $\left(t-n\tau\right)$, where the  time delay index $n$ ranges between 1 and 40. This results in a feature space dimension $D=120$ (see Fig.~2).      


\paragraph{Application of EOMR to mHW output data} 
In each of the forcing regimes we deploy the cross-validation procedure described above, and train EOMR models with hyperparameters sampled randomly from the following ranges of values: $d$ from $\left\{1,2,3\right\}$,  $K$ from $\in\left\{1,2,3\right\}$,  $\epsilon_R$ from the interval $\left[1,10\right]$,  $\epsilon_\gamma$ from the interval $\left[10^{-8},10^{-3}\right]$,  $\epsilon_W$ from the interval $\left[10^{-10},10^{-5}\right]$. 
 
 Results of the analysis are summarized in the Fig.~2.  As in the previous L96-example, despite of the low RMSE on the training data (left panel of Fig.~2A), both Deep Learning NN and RF/XGB perform worse then the trivial "persistent" predictor on the test data (right panel of Fig.~2A). Also TabPFN behaves similarly: being skilful on the training data, it  performs worse than the "persistent" predictor on the test data. Such a big discrepancy between the train and test performances is an indication of the overfitting phenomenon, and is also confirmed when comparing resulting model complexities to the training data size (Fig.~2C).  Deep Learning NN, RF/XGB and TabPFN v3.0 have around the same amount or more tuneable model parameters than the train data size, meaning that these models were "memorizing" and not learning.  Among the common tools only lasso MLR performs with RMSE of $1.5\times10^{-3}$, around 2 times smaller than the RMSE of the "persistent" predictor (with RMSE of $4\times10^{-3}$), and identifies a sparse model with around 122 tuneable regression parameters (see Fig.~2D), having RMSE performances on train and test data being close together.
 
 Optimal EOMR model determined from the cross-validation procedure from Sec.~\ref{sec:hyper} appears to have $d=1$, $K=1$, $\epsilon_R=5$, $\epsilon_W=8\cdot 10^{-9}$, $\epsilon_\gamma=0$. It results in the RMSE of around $4.5\cdot 10^{-4}$, almost an order of magnitude better than the "persistent" predictor. EOMR performances on the training and the validation data are very close together (Fig.~2A), and EOMR results in more than an order of magnitude sparser model then lasso MLR, with only around 8 tuneable parameters (compare Fig.~2D and Fig.~2E). Moreover, this very simple entropy-optimal model description for predictions of the dominant PCA mode of HsW that is "distilled" by EOMR, requires only the lag delayed values of the same dominant PCA mode from up to lag depth 9 - and all other features from the other PCA modes are redundant  (see Fig.~2E). In another words, EOMR has found a linear AutoRegressive Moving Average model (ARMA) to be the entropy-optimal regression approximation of the given time series data. This gives us an opportunity to deploy the very well established mathematical theory of ARMA time series processes, for example,  the Theorem 3.1.1 on page 83 from \cite{brockwell91}. According to this Theorem, ARMA model is weakly-stationary and causal if the roots of the respective characteristic polynomial are strictly outside of the unit circle in a complex plane. As can be seen from the Fig.~2F, all of the roots are strictly outside the unit circle - since all of their absolute values a strictly-larger than 1.0 (the smallest absolute value of a root is $1.015$). Hence, the model identified by EOMR is weakly-stationary and causal according to the Theorem, which means that this process will also be asymptotically-stable - and can be used as a stand-alone  "Ersatzmodel" for predictions and long-term simulations. 
 
 Computing predictions for this EOMR model requires only 15 elementary operations (8 additions and 7 multiplications, no divisions), and takes around $2\cdot 10^{-9}$ seconds - or 2 nanoseconds - on a common Mac LapTop. This is 160 times shorter time than the actual physical lag of  $0.316$ microseconds for the actual physical plasma process that we aim to predict here. In contrast, computing a single prediction instance with LLM-like TabPFN v3.0 model requires around  $2\cdot 10^{-2}$ seconds on the same LapTop, i.e., it is around five orders of magnitude slower then the actual lag of the predicted process.

\section{Discussion}
\label{sec:discuss}

Motivated by the universal approximation theorems, state-of-the-art AI currently follows a track guided by the so-called "double-descent principle", which states that as a machine learning model's capacity increases, its prediction error initially follows a classic U-shaped curve, spikes at the "interpolation threshold" where it perfectly fits (or "memorizes") the training data, and then decreases a second time as it enters a highly overparametrized regime \cite{belkin19}. The downside of this evolution are data-hungry and extremely-large LLM models that tend to memorize - and not to think and to learn \cite{shojaee25}, with up to tens of trillions of tuneable parameters,  and energy- as well as a money-consumption of a whole developed country. In the two  challenging examples from chaotic dynamical systems and fluid mechanics provided above, we illustrated the downsides of this evolution: even setting aside the energy consumption and complexity issues, such models (i) consistently exhibited overfitting behaviour (when perfect performance on the training and validation data is followed by a very poor test data performance, being inferior even to such dummy-models as the "persistent" predictor); and (ii) that the very high complexity of state-of-the-art models - and, particularly, the quadratic scaling of LLMs wrt. dimension $D$ and sample size $T$ - lead to extremely long computation times. For example, TabPFN transformer-based model required around $0.02$ seconds for predicting a single instance of the dominant PCA mode from the modified Hasegawa-Wakatani model, for a physical prediction lead time of $0.316$ microseconds, i.e., around 5-6 orders of magnitude slower than what would have been required for the online prediction of this system. 

Entropic AI tools  \cite{Horenko_2020,espa_22,horenko_pnas_23,bassetti25,groom26a,groom26b,horenko26,horenko26b} follow a paradigm that is orthogonal to the mainstream, and aim at finding models that are simultaneously as good as possible (in terms of performance), as well as are as simple and as unbiased as possible (in terms of information theory and Shannon entropy). In this manuscript, EOMR was proposed as a supervised extension of the recently-developed unsupervised EOMC, learning functional relations between features and targets by means of the low-dimensional piecewise-linear regression splines (\ref{eq:mod}). It was shown that the proposed EOMR approach has three major advantages: (i) it allows a \emph{simultaneous} joint learning of low-dimensional  subsets {\bf and} subspaces of relevant features; (ii) it was shown that the iteration computational and memory costs of EOMR scale linearly in $D$ and $T$; and (iii) formulated in the probabilistic way with entropic regularizations, it aims at finding sparse entropy-optimal models (\ref{eq:mod}). 

Surprisingly - or, may be, not surprisingly - dependent on the personal background and perspective, EOMR was shown to be a real game-changer in both of the provided examples. It resulted in the very sparse models (see Figs.~1 and 2), that are simultaneously orders of magnitude more performant and more simple than the considered state-of-the-art AI/ML models. Evaluation of predictions for EOMR model in the Hasegawa-Wakatani example required only 15 elementary operations (8 additions and 7 multiplications, no divisions), and took only around $2\cdot 10^{-9}$ seconds - or 2 nanoseconds - on a common Mac LapTop. This is 160 times shorter time than the actual physical lag of  $0.316$ microseconds for the underlying physical plasma process to be predicted - and, this performance can for sure be boosted even further on a more specialized hardware. This would open new possibilities for better online predictions, optimization and control of such complex, extremely-fast and chaotic systems.           

\bmhead{Availability of code}  Code and data can be uploaded from \url{https://seafile.rlp.net/d/4376396951164ef8ac78/}. 

\bmhead{Acknowledgement} The author would like to thank Davide Bassetti and Tim Prokosch (both RPTU Kaiserslautern-Landau), Rupert Klein (FU Berlin), Lukas Pospisil (VTU Ostrava), Michael Groom and Terry O'Kane (both CSIRO), and the other members of entropic AI community, discussions with whom provided a lot of motivation for this work. 
This work was funded by the EU Horizons project $AI4LUNGS$ (Grant Agreement No. 101080756).

\bibliography{MAD}

\end{document}